\documentclass{article}
\usepackage{spconf,amsmath,graphicx}
\usepackage{algorithmic}
\pdfoutput=1
\usepackage{booktabs}
\usepackage{amsmath}
\usepackage{setspace}
\usepackage{array,caption}
\usepackage[labelfont=bf]{caption}
\usepackage{CJKutf8}
\usepackage{amssymb}
\usepackage{tablefootnote}
\usepackage{caption}
\usepackage{subcaption}

\def\L{{\cal L}}

\title{Self-attention for Incomplete Utterance Rewriting}
\name{Yong Zhang, Zhitao Li, Jianzong Wang*, Ning Cheng, Jing Xiao\thanks{*Corresponding author: Jianzong Wang, jzwang@188.com. This paper is supported by the Key Research and Development Program of Guangdong Province  under grant No. 2021B0101400003 and the National Key Research and Development Program of China under grant No. 2018YFB0204403.}}
\address{Ping An Technology (Shenzhen) Co., Ltd.}
\begin{document}

%
\maketitle
\begin{abstract}
Incomplete utterance rewriting (IUR) has recently become an essential task in NLP, aiming to complement the incomplete utterance with sufficient context information for comprehension. In this paper, we propose a novel method by directly extracting the coreference and omission relationship from the self-attention weight matrix of the transformer instead of word embeddings and edit the original text accordingly to generate the complete utterance. Benefiting from the rich information in the self-attention weight matrix, our method achieved competitive results on public IUR datasets.
\end{abstract}
\vspace{-2.5mm}
\section{Introduction}
\vspace{-2.5mm}
\label{sec:intro}

The incomplete utterance rewriting (IUR)  has attracted dramatic attention in recent years due to its potential commercial value in conversation tasks. The main goal of IUR is to tackle the coreference and complement the ellipsis in the incomplete utterance and make the semantic information complete for understanding without referring to the context utterance. For the example of the multi-turn dialogue utterances $(u_1, u_2, u_3)$ in Table 1, $u_3$ is the incomplete utterance that omits the subject ``Shenzhen" and ``this" actually refers to the ``raining heavily recently" given the context utterances $u_1$ and $u_2$.

\begin{table}[h]
\begin{CJK*}{UTF8}{gbsn}
\vspace{-3mm}
\caption{The example of incomplete utterance rewriting}
\vspace{-3mm}

\scalebox{0.94}{
\begin{tabular}{c|c}
\hline Turns & Utterance (Translation) \\
\hline $u_1$ & 深圳的天气怎么样 \\
& (How is the weather in Shenzhen) \\
$u_2$ & 最近一直下暴雨 \\
& (It keeps raining heavily recently) \\
$u_3$ & 为什么这样 \\
& (why is this) \\
$u_3^*$ & 深圳为什么最近一直下暴雨 \\
& (Why is Shenzhen keeps raining heavily recently)
\end{tabular}
}
\footnotesize{ Notes: $u_1$ and $u_2$ denote the context
utterances in the dialogue and $u_3$ is the incomplete utterance with $u_3^*$ indicates the referenced complete utterance.}\\
\end{CJK*}
\vspace{-8mm}
\end{table}

\begin{figure*}[!tp]
\begin{center}
\includegraphics[width=0.9\textwidth]{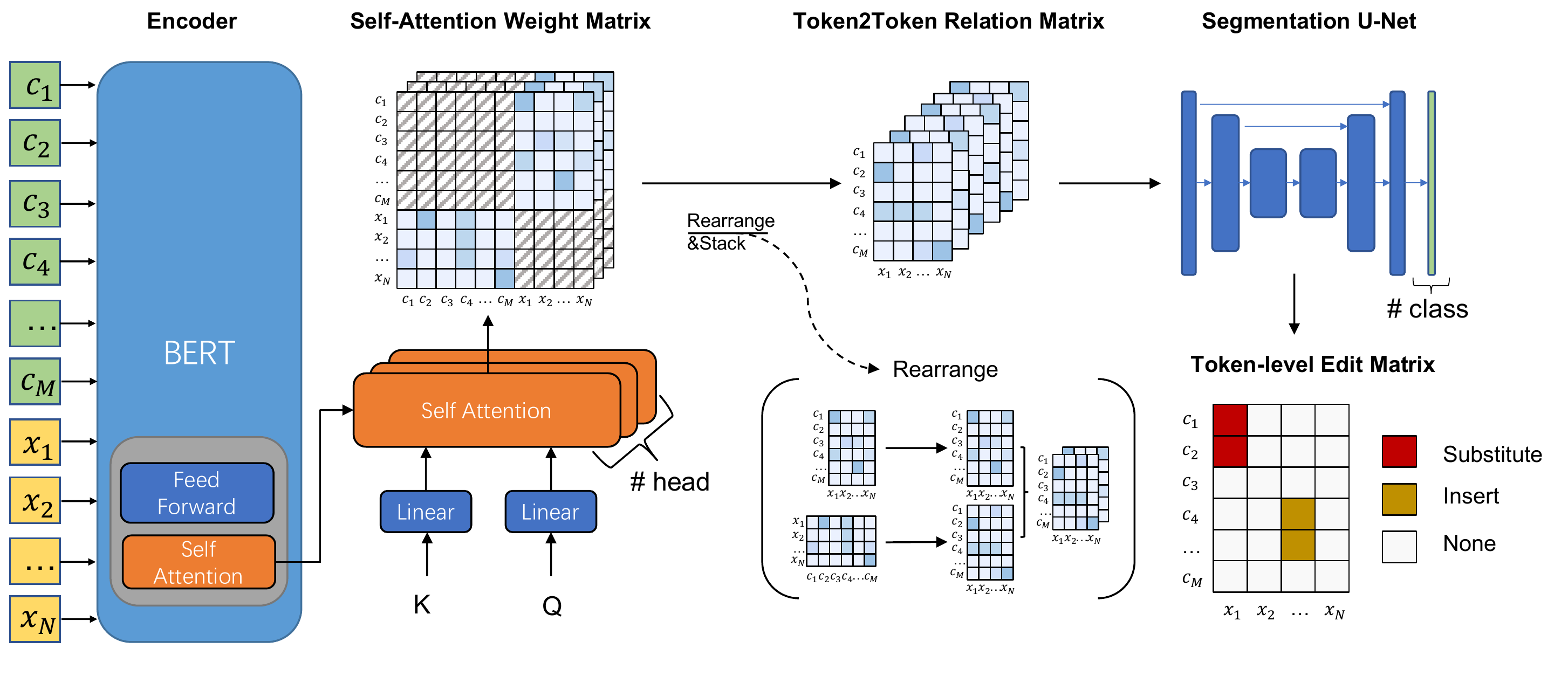}
\end{center}
\vspace{-8mm}
\caption{The architecture of our proposed model}
\label{run}
\vspace{-6.5mm}
\end{figure*}

Given most omitted and coreference words come from contexts utterances, current methods mainly apply the seq2seq methods with copy mechanism\cite{gu-etal-2016-incorporating}\cite{see2017get} or pointer network\cite{vinyals2015pointer} to deal with IUR. Su et al.\cite{su2019improving} proposes a hyper-parameter $\lambda$ to distinguish the attention of context and incomplete utterance based on transformer-based seq2seq model and pointer network. Pan et al.\cite{pan2019improving} apply a ``pick and combine" (PAC) method, which first picks omitted words in the context utterances and utilizes the pointer generative network to take omitted words as extra features to produce the output. CSRL\cite{xu2020semantic} exploits additional information from semantic role labeling (SRL) to enhance BERT representation for rewriting utterances, which requires more processes. 
Although they achieved promising results, they still unavoidably suffer from exposure bias and low autoregressive generation speed. \par 
To improve the speed, SARG\cite{huang2021sarg} fuses the sequence labeling and non-autoregressive generation first to identify required operations for incomplete utterance and insert words from context utterances to the incomplete utterance accordingly. RAST\cite{hao2020robust} formulates IUR task as a span prediction task of deletion and insertion with reinforcement learning to improve fluency. RUN\cite{liu2020incomplete} formulates the IUR task as semantic segmentation based on the feature map constructed by the similarity function on the word embeddings and achieves better performance with faster speed.
\par 
Above mentioned methods depend heavily on encoders’ output which could be the information bottleneck whereas rich semantics dependency information hidden in the attention weight matrix was overlooked. In this work, we propose to shed more light on the signal hidden in the self-attention weight matrix and leverage a segmentation CNN from computer vision to extract more information for the IUR task. The self-attention weight matrix can naturally capture the coreference and omission relationships between the context utterances and the incomplete utterance. Without outputting the word embedding, we directly apply a segmentation CNN to map the learned token2token relationship in the self-attention weight matrix to edit operations in parallel. The final complete utterance can be produced by editing the incomplete utterance and context utterances based on the generated edit type tags. 
Our contributions are summarized below:\\
 
\vspace{-5.4mm}
\begin{enumerate}
  \item We explore the self-attention weight for the token relationship representation and apply it to the IUR.
  \vspace{-0.9mm}
  \item We propose a straightforward and efficient model structure based on the self-attention weight matrix with low resource cost.
  \item Experimental results demonstrate that our proposed method performs better than current baselines on the RESTORATION\cite{pan2019improving} and REWRITE\cite{su2019improving} benchmark.
\end{enumerate}

\vspace{-7mm}
\section{Proposed Method}
\label{sec:method}
\vspace{-2.5mm}

In this section, we will introduce our method in detail. As shown in Figure 1, we propose a straightforward model structure with BERT\cite{devlin2019bert} as the encoder to produce the token2token relation matrix and U-Net\cite{ronneberger2015u} as the classifier. We named our model as Rewritten Attention U-Net (RAU).

Formally, given multi-turn dialogue utterances $(u_1,u_2,$ $…,u_t)$, we concatenate all context utterances $(u_1,u_2,…,$ $u_{t-1})$ into an M-length word sequence $c = (c_1,c_2,…,c_M)$ and using a special mask $[SEP]$ to separate different utterance. Besides, the last utterance in the dialogue: the incomplete utterance $u_t$ is denoted as an N-length word sequence $x = (x_1,x_2,…,x_N)$.

\vspace{-4mm}
\subsection{Token2Token Relation Mapping}
\vspace{-1.5mm}
\label{ssec:subhead}

{\bf Encoder } 
We use a pre-trained language model BERT\cite{devlin2019bert} as the encoder to learn the context information. The concatenation of context utterance sequence c and incomplete utterance sequence $x$ will first be passed to the corresponding tokenizer to generate tokens and further processed by the BERT to get the contextual information among utterances. Since the model does not require the hidden state of the word for representation, the last layer's feed-forward layer is abandoned in the structure. 

\noindent{\bf Token2Token Relation Matrix } 
On top of the context-aware information learned by BERT, we propose to apply BERT's self-attention weight matrix as the representation of the classifier to learn edit operations. With pre-trained knowledge, the self-attention weight of each layer can further learn the token to token positional, syntax, and semantic relationship. And different heads of the layer pays attention to diverse perspective. 

The calculation of self-attention weight\cite{vaswani2017attention} relies on the query dimensionality $d_{q} $ and the key dimensionality $d_{k}$. For each token, dot products are performed by the query with all keys among the tokens in the input and divided each by $\sqrt{d_{k}}$ to smooth the value. Finally, a softmax function is applied to obtain the attention weights distribution. And the attention weight can be calculated simultaneously by packing queries and keys together into matrix $Q$ and matrix $K$ as:

\vspace{-6mm}
\begin{equation}
\text { Attention Weight }(Q, K)=\operatorname{softmax}\left(\frac{Q K^{T}}{\sqrt{d_{k}}}\right)
\vspace{-2mm}
\end{equation}

Multi-head attention allows the model to learn the information from different aspects with different sets of query weight matrixes $W_{i}^{Q}$ and key weight matrixes $W_{i}^{K}$. $Head_i \in \mathbb{R}^{(M+N) \times (M+N)}$ is self-attention weight matrix with ${i}$ indicates the corresponding head.

\vspace{-6.7mm}
\begin{equation}
\text { Head }_{i}=\text { Attention Weight}\left(Q W_{i}^{Q}, K W_{i}^{K}\right)
\vspace{-2.7mm}
\end{equation}

Since the self-attention weight matrix includes the self-dependency of each token, the model has to select desired attention of the token in context utterances with the token in the incomplete utterance. As shown in the Token2Token Relation Matrix of Figure 1, for each head's self-attention weight matrix, the top right and the bottom-left part corresponding to the token relationship between the context utterance and the incomplete utterance are selected. And rearrange is required for the bottom left part to maintain the same shape and the order of the token. Finally, for each attention head, it can acquire a token2token relation weight matrix $\text { Head }_{i}^{*} \in \mathbb{R}^{M \times N \times 2}$:
\vspace{-6.7mm}


\begin{equation}
\text { Head }_{i}^{*}=\text { Slice }_{1}\text { Head }_{i} \oplus \text { Rearrange }(\text { Slice }_{2}\text { Head }_{i})
\end{equation}

Where $Slice_1$ and $Slice_2$ respectively corresponds to the mentioned two selection operations and $\oplus$ indicates the concatenation.

\noindent{\bf Visualization } 
As mentioned before, self-attention with different heads can help recognize the position, syntax, and semantic information. We statistically analyze the last layer's self-attention weight matrix to complement the proposed method. As shown in Figure 2, it can be observed that most of the heads of the last layer pay more attention to semantic information (Coreference and Omission). Also, different heads will learn some syntax relationships and other information. Take Figure 3's one head's self-attention weight matrix visualization\cite{su2019improving} as an example, this head has aligned the coreference subject ``My daughter" in $c$ with the pronoun ``She" in the $x$, representing the semantic ability. Besides, it also highlights the omission of ``eat napkins" in the target insertion position. We argue it is due to the head's position detection ability to identify the position of the current token in the correct word order cooperated with semantic knowledge. 

\vspace{-1mm}
\begin{figure}[!htbp]
\vspace{-3mm}
\includegraphics[width=0.46\textwidth,height=0.30\textwidth]{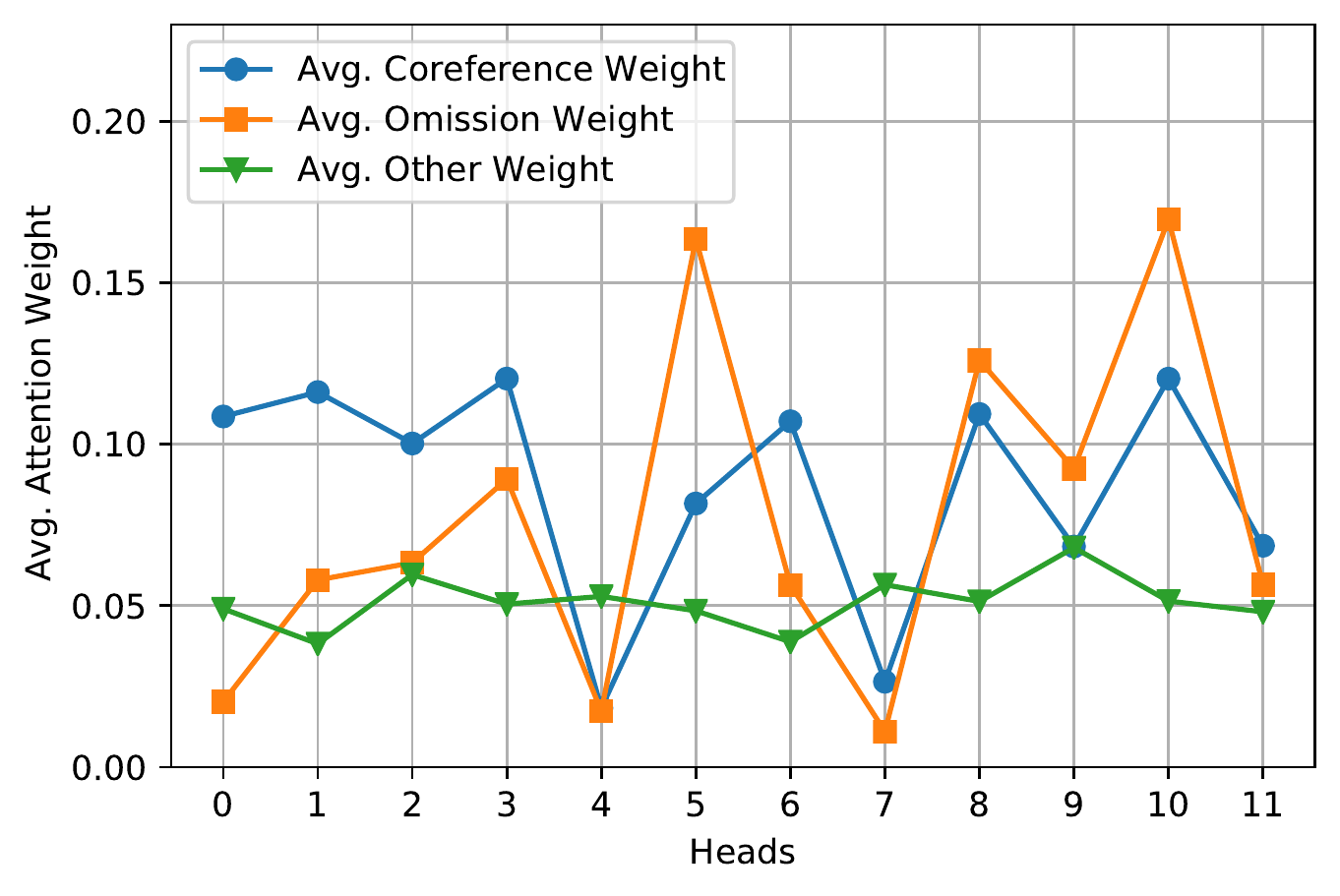}
\vspace{-4.7mm}
\caption{Self-attention weight statistic of BERT last layer's 12 heads on the REWRITE dev set. Coreference and Omission: mentioned IUR token2token relationship. Other: other relationship. Avg. Weight: the average weight of all heads's attention for the same token2token relation type in all relation matrix cells.}
\label{fig}
\vspace{-4.5mm}
\end{figure}

Another advantage is that utilizing the self-attention weight matrix could simplify the model architecture omitting the feedforward structure of the last layer and contribute to the speed of the training and prediction.

\begin{figure}[!htbp]
\vspace{-10mm}
\includegraphics[width=0.5\textwidth,height=0.6\textwidth]{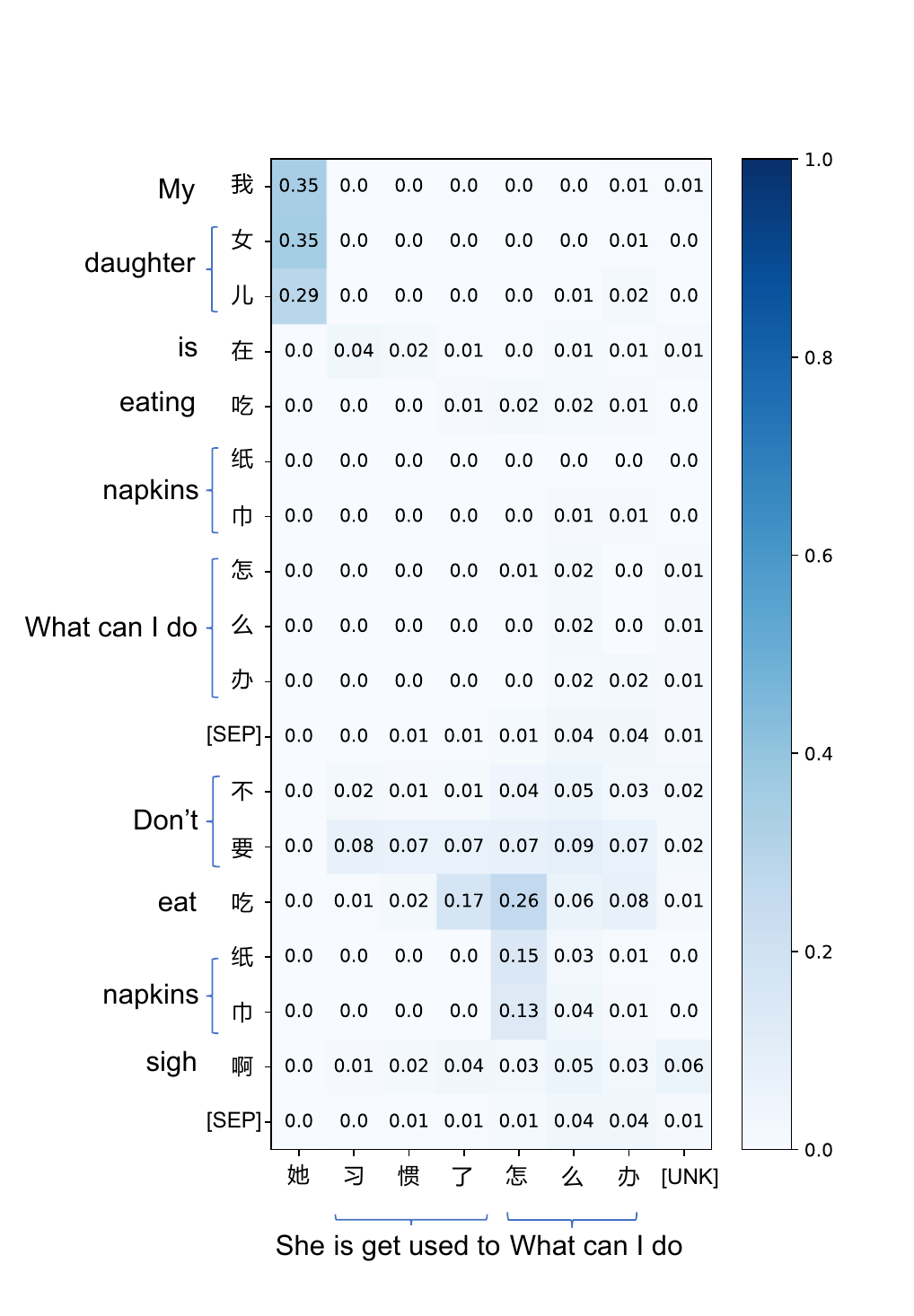}
\vspace{-9.8mm}
\caption{Example of BERT last layer's one head's self-attention weight matrix}
\label{fig}
\vspace{-7mm}
\end{figure}

\noindent{\bf Segmentation } 
Regarding the token2token relation matrix as a multi-channel image, we apply the U-Net\cite{ronneberger2015u} to integrate low-level features to high-level features and as the classifier to map the token2token relationship to the corresponding edit type. U-Net is proposed for image segmentation in the area of CV, and it is originally used for pixels' interactively parallel classification, which is naturally suitable in our case. The down-sampling blocks of U-Net can enlarge the receptive field of token-to-token relevance embedding$\text{ Head }_{i}^{*}\left(c_m,x_n\right)$ and fuse the global information learned from the encoder. And the up-sampling blocks help distribute the fused information to each cell. The output of U-Net is the same width and height as the input matrix with channel amount aligned with the edit operation amount (Substitute, Insert, and None). Each cell of the channel matrix corresponds to the score of the edit type.\vspace{-3mm}

\vspace{-2mm}
\begin{equation}
\vspace{-2mm}
F=\text { U-Net }( \oplus_i^I (\text { Head }_{i}^*))
\end{equation}
\begin{equation}
Edit\left(c_m,x_n\right) = \text { ArgMax }{F}\left(c_m,x_n\right)
\end{equation}
\vspace{-6mm}

where $I$ is the amount of heads and $\oplus_i^I (\text { Head }_{i}^*) $ denotes concatenating all ${ Head }_{i}^*$. The $F \in \mathbb{R}^{M \times N \times C}$ is the output of the U-Net with $C$ class channels. The class of each cell $Edit\left(c_m,x_n\right)$ is the $\text {ArgMax }$ of $ {F}\left(c_m,x_n\right) \in \mathbb{R}^{1 \times 1 \times C}$.

\vspace{-4.2mm}
\subsection{Incomplete Utterance Edit}
\label{ssec:subhead}
\vspace{-2mm}

After obtaining the token-level editing matrix $Edit \in \mathbb{R}^{M \times N}$ with each entry of the matrix represents the token2token editing type between $c$ and $ x$, we can use a simple editing algorithm to generate the complete utterance. The example is shown in Figure 3, the coreference relationship $\rightarrow$ Substitute operation: ``My daughter" will substitute the ``She" in $x$, and the omission relationship $\rightarrow$ Insert before operation: ``eat napkins" will be inserted before the ``What can I do". Nothing is done for None operation of the other relationship.

\begin{center}
\begin{table*}
\caption{The results of all compared models trained and evaluated on the RESTORATION.}
\vspace{-2mm}
\scalebox{0.96}{
\begin{tabular}[Ht]{lccccccccccccc}
\hline Model & $\mathcal{P}_{1}$ & $\mathcal{R}_{1}$ & $\mathcal{F}_{1}$ & $\mathcal{P}_{2}$ & $\mathcal{R}_{2}$ & $\mathcal{F}_{2}$ & $\mathcal{P}_{3}$ & $\mathcal{R}_{3}$ & $\mathcal{F}_{3}$ & $\mathbf{B}_{1}$ & $\mathbf{B}_{2}$ & $\mathbf{R}_{1}$ & $\mathbf{R}_{2}$ \\
\hline 
T-Ptr-$\lambda$\cite{su2019improving} & $-$ & $-$ & $51.0$ & $-$ & $-$ & $40.4$ & $-$ & $-$ & $33.3$ & $90.3$ & $87.4$ & $90.1$ & $83.0$ \\
PAC\cite{pan2019improving} & $70.5$ & $58.1$ & $63.7$ & $55.4$ & $45.1$ & $49.7$ & $45.2$ & $36.6$ & $40.4$ & $89.9$ & $86.3$ & $91.6$ & $82.8$ \\
CSRL\cite{xu2020semantic} & $-$ & $-$ & $-$ & $-$ & $-$ & $-$ & $-$ & $-$ & $-$ & $90.6$ & $89.7$ & $91.1$ & $85.0$  \\
SARG\cite{huang2021sarg} & $-$ & $-$ & $62.4$ & $-$ & $-$ & $52.5$ & $-$ & $-$ & $46.3$ & $92.2$ & $89.6$ & $92.1$ & $\mathbf{86.0}$ \\
RAST\cite{hao2020robust} & $-$ & $-$ & $-$ & $-$ & $-$ & $-$ & $-$ & $-$ & $-$ & $90.4$ & $89.6$ & $91.2$ & $84.3$ \\
RUN (BERT)\cite{liu2020incomplete}& $7 3 . 2$ & $64.6$ & $68.6$ & $5 9 . 5$ & $53.0$ & $56.0$ & $50.7$ & $45.1$ & $47.7$ & $92.3$ & $8 9 . 6$ & $9 2 . 4$ & $85.1$ \\
RAU (Ours) & $\mathbf{75.0}$ & $\mathbf{65.5}$ & $\mathbf{69.9}$ & $\mathbf{61.2}$ & $\mathbf{54.3}$ &      $\mathbf{57.5}$ & $\mathbf{52.5}$ & $\mathbf{47.0}$ & $\mathbf{49.6}$ & $\mathbf{92.4}^{*}$ & $\mathbf{89.6}$ & $\mathbf{92.8}^{*}$ & $\mathbf{86.0}^{*}$ \\
\hline
\end{tabular}
}
\footnotesize{ Notes: $P_n$, $R_n$, and $F_n$ denote precision, recall, and F-score of n-grams restored word in rewritten utterance based on incomplete and complete utterances. The detail is described in restoration score\cite{pan2019improving}. $B_n$ indicates n-gram BLEU score and $R_n$ represents n-gram ROUGE score. The - indicates result is not reported in the paper. And the $^{*}$ indicates the result is statistically significant against all the baselines with the p-value ${<}$ 0.05. The marks are the same for Table 3.} \\

\vspace{-9.5mm}
\end{table*}
\end{center}

\vspace{-16.5mm}
\section{Experiments}
\label{sec:Experiments}

\vspace{-3.5mm}
\subsection{Setup}
\label{ssec:subhead}
\vspace{-2mm}

{\bf Datasets } 
 We conduct our experiments on RESTORATION-200K \cite{pan2019improving} and REWRITE\cite{su2019improving} which are split as $0.8/0.1/0.1$ and $0.9/0.1/-$ for training/development/testing according to the previous methods. The dataset consists of multi-turn dialogue sentences as input and ``correctly" rewritten sentences as label.


\noindent{\bf Comparing methods }
We compare the performance of our method with the following methods as described in INTRODUCTION: the transformer based pointer generator (T-Ptr-Gen)\cite{see2017get}, T-Ptr-$\lambda$\cite{su2019improving}, PAC\cite{pan2019improving}, CSRL\cite{xu2020semantic}, SARG\cite{huang2021sarg}, RAST\cite{hao2020robust}, and RUN (BERT)\cite{liu2020incomplete}. For benchmark details, please refer to the corresponding paper.

\noindent{\bf Evaluation }
We follow the previous work's usage to apply BLEU\cite{papineni2002bleu}, ROUGE\cite{lin-2004-rouge}, EM and restoration score\cite{pan2019improving} as the automatic evaluation metrics to compare our proposed method with others.

\noindent{\bf Model setting }
We utilize bert-base-chinese from HuggingFace’s community\cite{wolf2020transformers} as our pre-trained BERT and it is fine-tuned as part of the training. The number of layers is 12 with 12 attention heads. Only the last layer's self-attention weight is used since it achieves the best result in our experiment. Adam\cite{KingmaB14} is utilized to optimize the model with a learning rate of 1e-5. Weighted cross-entropy is applied to address the imbalanced class distribution of mentioned three edit operations.

\begin{table}[!htbp]
\centering
\vspace{-3mm}
\caption{The results of all compared models trained and evaluated on the REWRITE. }
\vspace{-2mm}
\scalebox{0.88}{
\begin{tabular}{lccccc}
\hline Model & $\mathbf{E M}$ & $\mathbf{B}_{2}$ & $\mathbf{B}_{4}$ & $\mathbf{R}_{2}$ & $\mathbf{R}_{L}$ \\
\hline 
T-Ptr-Gen\cite{see2017get} & $53.1$ & $84.4$ & $77.6$ & ${85.0}$ & $89.1$ \\
T-Ptr-$\lambda\cite{su2019improving}$ & $52.6$ & $85.6$ & $78.1$ & ${85.0}$ & $89.0$ \\
T-Ptr-$\lambda$ (BERT)\cite{su2019improving} & $57.5$ & $86.5$ & $79.9$ & $86.9$ & $90.5$ \\
RUN (BERT)\cite{liu2020incomplete} & ${6 6 . 4}$ & ${9 1 . 4}$ & ${8 6 . 2}$ & ${9 0 . 4}$ & ${9 3 . 5}$ \\
RAU (Ours) & $\mathbf{6 8 . 4}^{*}$ & $\mathbf{9 1 . 6}^{*}$ & $\mathbf{8 6 . 6}^{*}$ & $\mathbf{9 0 . 6}^{*}$ & $\mathbf{93.9}^{*}$ \\
\hline
\end{tabular}
}
\footnotesize{ Notes: $EM$ indicates the exact match score and $R_L$ is ROUGE score based on the longest common subsequence (LCS).}\\
\end{table}
\vspace{-10.2mm}

\subsection{Main Result}
\vspace{-2mm}
\label{ssec:subhead}

The result of Restoration and Rewrite are shown in Table 2 and Table 3.  For Restoration, our method performs better than the previous best model RUN (BERT) in all n-grams F-score, that  $P_1$, $P_2$ and $P_3$ averagely raise 2.1 points and different n-grams recall achieves comparable performance. The result indicates our method can help correctly recognize more target words with the help of sufficient information of attention weight. In addition, our model outperforms the previous model on all the BLEU and ROUGE. Although the improvement is slight, it also supports our model is robust since the BLEU and ROUGE scores of all previous models are close even restoration scores are different and our model has the highest restoration score.

For Rewrite, our method also performs better on all scores, significantly improves 2 points on the most challenging EM score, which requires an exact match of rewritten utterance with the referenced complete utterance.

\vspace{-4.5mm}
\subsection{Ablation Study}
\vspace{-2mm}
\label{ssec:subhead}
We conduct a series of ablation studies to evaluate the effectiveness of attention weight learned by different layers and heads of BERT. The results are depicted in Table 4.

\begin{table}[!htbp]
\caption{Ablation results on the RESTORATION test set. } 
\vspace{-2mm}
\scalebox{0.95}{
\begin{tabular}{lccccc}
\hline Model & $\mathcal{F}_{1}$ & $\mathcal{F}_{2}$ & $\mathcal{F}_{3}$ & $\mathbf{B}_{2}$ & $\mathbf{R}_{2}$ \\
\hline RAU $L12$ & $69.9$ & $57.5$ & $49.6$ & $\mathbf{89.6}$ & $\mathbf{8 6 . 0}$ \\
RAU $L6$ & $69.9$ & $55.3$ & $46.3$ & $86.2$ & $84.1$ \\
RAU $L1$ & $58.8$ & $44.3$ & $35.2$ & $83.8$ & $80.9$ \\
RAU $L6,12$ & $\mathbf{70.8}$ & $57.5$ & $49.1$ & $87.2$ & $84.9$ \\
RAU $L1,6,12$ & $70.7$ & $\mathbf{58.0}$ & $\mathbf{5 0 . 0}$ & $87.0$ & $85.0$ \\
RAU $L$ $all$ & $70.2$ & $57.0$ & $48.5$ &$87.9$ & $85.1$ \\
RAU $L12$ $H1-6$ & $70.2$ & $57.1$ & $48.7$ & $89.2$ & $85.6$ \\
\hline
\end{tabular}
}
\footnotesize{ Notes: L and H denote the layer and head of BERT with the next digit indicates the index from 1-12; ``L all" means all layers are included. }\\
\vspace{-7mm}
\end{table}


As expected, the higher the layer, the better high-level information can be learned by the head attention. All evaluation metric scores drop consistently with lowering the layer. Given the phenomenon observed by Jawahar et al.\cite{jawahar2019does} that the lower layer tends to learn the surface feature, the middle and the high layer prefer the syntax feature and semantic feature, we also try to aggregate different layer's attention into the token2token matrix. All combination's experiment result indicates last layer's information is far sufficient for the current task. We also observe that learned different level information distributes in various heads of different layers, and some heads may be lazy, which is similar to the previous observation\cite{clark2019does}. We try to prune the heads with the first six kept. The result shows that BERT can transfer the learned information to desired heads with finetuning setting.

\vspace{-4.2mm}
\section{Conclusions}
\label{sec:conlusion and future work}
\vspace{-2.5mm}

In this paper, we discovered the potential usage of the overlooked self-attention weight matrix from the transformer and proposed a straightforward and effective model for the IUR task. Our model has achieved state-of-the-art performance on public IUR datasets. Deeper research on the incorporation of self-attention weight matrix for other NLP tasks and linguistics studies can be conducted in the future.

\vspace{-4.2mm}
\section{ACKNOWLEDGEMENTS}
\label{sec:ACKNOWLEDGEMENTS}
\vspace{-2.5mm}

This paper is supported by the Key Research and Development Program of Guangdong Province  under grant No. 2021B0101400003 and the National Key Research and Development Program of China under grant No. 2018YFB0204403. Corresponding author is Jianzong Wang from Ping An Technology (Shenzhen) Co., Ltd (jzwang@188.com).

\bibliographystyle{IEEEbib}
\bibliography{strings,refs}

\end{document}